\documentclass[letterpaper, 10 pt, conference]{ieeeconf}  

\IEEEoverridecommandlockouts                              

\usepackage{graphics} 
\usepackage{epsfig} 
\usepackage{mathptmx} 
\usepackage{times} 
\usepackage{amsmath} 
\usepackage{amssymb}  
\usepackage{booktabs}
\usepackage{multirow}
\usepackage{cite}
\usepackage{algorithm}
\usepackage{algorithmic}

\title{\LARGE \bf
Closed-Loop Validation-Repair for Healthcare Interoperability: \\
A Multi-Model Study of Schema Compliance in Clinical LLMs
}

\author{Jianru Shen%
\thanks{Jianru Shen is with the University of Montana,
        Missoula, MT 59812, USA
        {\tt\small js258133@umconnect.umt.edu}}%
}

\begin{document}

\maketitle
\thispagestyle{empty}
\pagestyle{empty}

\begin{abstract}
Healthcare interoperability requires AI systems to produce structured outputs conforming to standardized schemas including ICD-10 for diagnostic coding, CPT for procedure billing, and HL7 FHIR for data exchange. While large language models demonstrate clinical reasoning capabilities, their integration into electronic health record systems faces a critical barrier: schema noncompliance. We evaluate three open-source models, Qwen2.5 7B, Llama 3.1 8B, and Gemma2 9B, via local deployment across 320 clinical scenarios spanning ten medical specialties, yielding 960 model--scenario pairs assessed under paired baseline and validation-repair conditions. First, schema noncompliance is consistent across the three model families, with baseline compliance rates ranging from 85.9 to 91.6 percent despite varying architectures and training data, suggesting shared gaps in medical training corpora rather than model-specific limitations. Second, 96 percent of validator-detected failures are representation-level format violations such as alternative medical abbreviations and code prefixes, indicating models follow clinical writing conventions but lack awareness of healthcare IT standards. Third, the validation-repair framework achieves 99.0 percent overall compliance, ranging from 98.4 to 99.4 percent across models, with most errors resolving within one or two iterations. Exact McNemar p-values below 0.001 and absolute improvements of 7.8 to 12.5 percentage points across model sizes confirm statistical significance. These results support closed-loop validation-repair as an effective system-level safeguard for healthcare interoperability, improving schema-level readiness for downstream clinical system integration.
\end{abstract}

\section{INTRODUCTION}

Modern healthcare delivery relies on seamless interoperability among diverse information systems. Electronic health record platforms, built on standardized schemas including HL7 FHIR for resource representation~\cite{c4}, ICD-10 for diagnostic coding~\cite{c5}, and CPT for procedure billing~\cite{c6}, enable data exchange across institutions and clinical workflows. However, integrating AI-powered clinical decision support into these ecosystems presents a critical challenge. While large language models demonstrate expert-level clinical reasoning on medical examinations and diagnostic tasks~\cite{c1,c2,c3}, their outputs frequently violate schema constraints required for system interoperability.

We identify a systematic deployment barrier that transcends individual model architectures: clinical competence does not guarantee schema compliance. Models produce clinically valid outputs that violate healthcare IT standards, for example using the international abbreviation BD rather than the US standard BID for twice daily dosing. Such violations occur at rates of 8 to 14 percent and prevent EHR integration, where single format violations cause billing rejections or workflow disruptions.

This paper presents a multi-model empirical study addressing healthcare interoperability gaps in clinical LLMs. We investigate three research questions. First, is schema noncompliance architecture-agnostic across different LLM vendors and model sizes? Second, what types of errors cause schema violations and do they reflect medical knowledge gaps or format compliance issues? Third, can iterative validation-repair achieve deployment-oriented schema compliance exceeding 98 percent across diverse architectures?

\begin{figure*}[t]
\centering
\includegraphics[width=0.82\textwidth]{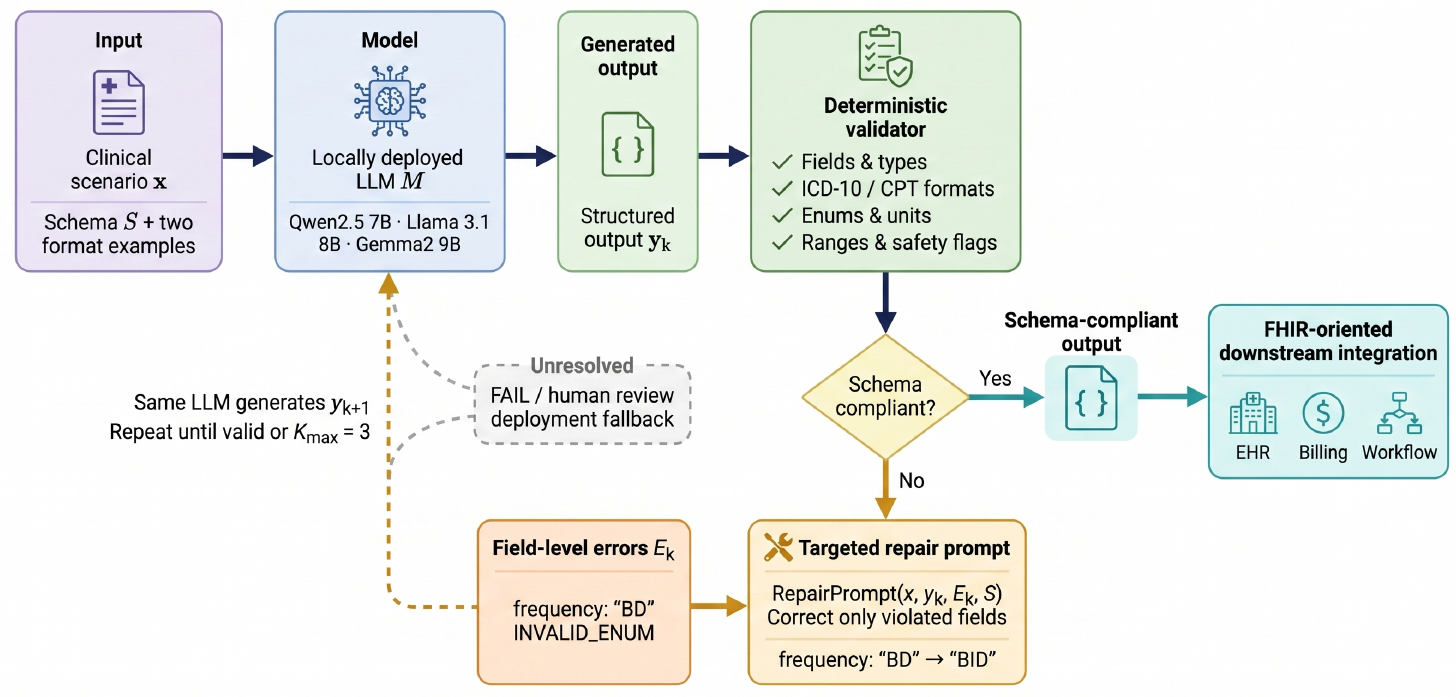}
\caption{Overview of the closed-loop validation-repair framework. A clinical scenario with schema specification and two format examples is processed by a locally deployed LLM; a deterministic validator checks fields, types, code formats, enumerations, ranges, and safety flags. Detected field-level errors are converted into a targeted repair prompt and the same model regenerates the output, repeating until compliance or $K_{\max}=3$; unresolved scenarios are recorded as failures with human review as the proposed deployment fallback.}
\label{fig:overview}
\end{figure*}

Our multi-model findings support schema validation-repair as an effective closed-loop system component, where validation provides feedback signals and repair executes corrective control, independent of advances in individual model architectures. Figure~\ref{fig:overview} provides an overview of the framework.

\section{RELATED WORK}

\subsection{LLMs in Clinical Applications}

LLMs demonstrate strong performance on medical question answering, diagnostic reasoning, and clinical text tasks~\cite{c1,c2,c3,c16}. However, these evaluations focus on clinical accuracy metrics such as exam scores and diagnostic correctness, assuming generated outputs conform to required formats rather than addressing the structured output compliance required for healthcare system integration.

\subsection{Healthcare Interoperability Standards}

Healthcare information exchange relies on standardized schemas: HL7 FHIR for resource-based clinical data representation~\cite{c4}, ICD-10 for diagnostic coding with structured alphanumeric formats~\cite{c5}, and CPT for procedure billing with strict five-digit numeric codes~\cite{c6}. Fragmented data standards have long been recognized as a barrier to health information exchange~\cite{c20}. Validation tools exist for human-authored data, but traditional validators report conformance errors and do not themselves generate corrected content~\cite{c29}. Recent work has applied LLMs to map clinical text to FHIR resources~\cite{c23} and has coupled an LLM with iterative FHIR-validator feedback for syntactic validity~\cite{c24}. Our study differs in evaluating three locally deployed open-weight models under a common study-defined schema, with paired compliance statistics and a cross-model error taxonomy.

\subsection{Structured Output Generation and Validation}

Constrained decoding and grammar-guided generation enforce syntactic validity by restricting outputs to permitted structures~\cite{c8,c9}, but do not address semantic compliance with domain-specific schemas. Grammar-constrained decoding extends such guarantees to a broad range of structured NLP tasks~\cite{c30}; in contrast, our framework applies deterministic post-generation validation and targeted regeneration, keeping domain-specific checks external to the decoding engine. Self-refinement techniques including Reflexion~\cite{c10} and Self-Refine~\cite{c11} iteratively improve outputs through self-critique, but target task accuracy rather than format compliance. Constitutional AI aligns model behavior with specified principles rather than data format constraints~\cite{c12}.

\subsection{Positioning and Contribution}

Prior healthcare LLM studies typically evaluate single models, limiting conclusions about whether compliance issues are model-specific or systematic. Our multi-model evaluation shows that schema noncompliance is consistent across three vendors, and that validation alone is insufficient without repair. We contribute a paired multi-model compliance evaluation, an error taxonomy showing failures are predominantly representation-level format violations, and a convergence analysis of closed-loop repair.

\section{VALIDATION-REPAIR FRAMEWORK}

We present a closed-loop validation-repair framework that addresses schema noncompliance through cybernetic error correction~\cite{c28}. The framework implements a feedback control architecture where validation acts as a compliance sensor detecting schema violations, and repair functions as a corrective controller generating schema-compliant outputs through targeted error feedback. This cybernetic approach enables vendor-independent compliance improvement while preserving the clinical reasoning capabilities of foundation models.

The framework consists of three components. First, we define healthcare interoperability requirements that LLM outputs must satisfy for downstream clinical system integration. Second, we describe the validation-repair algorithm that iteratively corrects schema violations. Third, we specify evaluation metrics that enable cross-model performance comparison and convergence analysis.

\subsection{Healthcare Interoperability Requirements} Healthcare systems require LLM outputs to conform to standardized schemas that enable interoperability across electronic health record platforms, billing systems, and clinical workflows. Core requirements include diagnostic codes following ICD-10 format specifications, procedure codes adhering to CPT standards, medication frequencies using standardized clinical abbreviations, and dosage units from predefined medical vocabularies.

Specifically, outputs must satisfy the following constraints: diagnostic codes must match the ICD-10 regex pattern matching alphabetic prefix followed by numeric codes with optional decimal separators; procedure codes must be five-digit numeric strings or recognized standard letter codes; medication frequencies must belong to the enumeration set consisting of QD, BID, TID, QID, Q4H, Q6H, and PRN; dosage units must be selected from mg, ml, g, mcg, or units; and confidence scores must fall within the zero to one range. Additionally, safety flags indicating allergy checking and drug interaction screening must be set to true. These constraints constitute a structured interoperability schema incorporating FHIR-oriented fields together with ICD-10 and CPT format requirements. Validation targets code-format validity rather than full FHIR profile conformance or terminology-level verification that a given code exists in the current code set; the official FHIR \$validate operation, by contrast, can assess resources against schemas, constraint rules, profiles, and terminology rules~\cite{c29}.

Noncompliance with these requirements prevents integration with EHR systems, introduces billing errors that cause claim rejections, and disrupts clinical workflows. A single format violation can render an otherwise clinically accurate decision unusable for downstream system integration.

\subsection{Validation-Repair Algorithm}

Our framework employs iterative error-aware refinement to achieve schema compliance. Algorithm~\ref{alg:repair} presents the complete procedure.

\begin{algorithm}[t]
\caption{Closed-Loop Validation-Repair}
\label{alg:repair}
\begin{algorithmic}[1]
\REQUIRE Clinical scenario $x$, LLM $M$, schema $S$, $K_{\max} = 3$
\ENSURE Schema-compliant decision $y$, or FAIL
\STATE $y_0 \leftarrow M(x)$ \COMMENT{initial generation}
\FOR{$k = 0$ \TO $K_{\max}$}
    \STATE $E_k \leftarrow \mathrm{Validate}(y_k, S)$ \COMMENT{feedback sensor}
    \IF{$E_k = \emptyset$}
        \RETURN $y_k$ \COMMENT{compliance achieved}
    \ENDIF
    \IF{$k = K_{\max}$}
        \STATE \textbf{break}
    \ENDIF
    \STATE $\mathit{prompt} \leftarrow \mathrm{RepairPrompt}(x, y_k, E_k, S)$ \COMMENT{controller}
    \STATE $y_{k+1} \leftarrow M(\mathit{prompt})$ \COMMENT{corrective action}
\ENDFOR
\RETURN FAIL
\end{algorithmic}
\end{algorithm}

\begin{algorithm}[t]
\caption{Deterministic Field-Level Schema Validation}
\label{alg:validate}
\begin{algorithmic}[1]
\REQUIRE Structured output $y$, schema $S$
\ENSURE Field-level error set $E$
\STATE $E \leftarrow \emptyset$
\FORALL{required fields $f$ in $S$}
    \IF{$f$ is absent from $y$}
        \STATE $E \leftarrow E \cup \{(f, \mathrm{MISSING\_FIELD})\}$; \textbf{continue}
    \ENDIF
    \IF{$\mathrm{Type}(y[f]) \neq S.\mathrm{type}[f]$}
        \STATE $E \leftarrow E \cup \{(f, \mathrm{TYPE\_MISMATCH})\}$; \textbf{continue}
    \ENDIF
    \IF{$S.\mathrm{enum}[f]$ is defined \AND $y[f] \notin S.\mathrm{enum}[f]$}
        \STATE $E \leftarrow E \cup \{(f, \mathrm{INVALID\_ENUM})\}$
    \ENDIF
    \IF{$S.\mathrm{pattern}[f]$ is defined \AND $y[f]$ does not match $S.\mathrm{pattern}[f]$}
        \STATE $E \leftarrow E \cup \{(f, \mathrm{INVALID\_FORMAT})\}$
    \ENDIF
    \IF{$S.\mathrm{range}[f]$ is defined \AND $y[f] \notin S.\mathrm{range}[f]$}
        \STATE $E \leftarrow E \cup \{(f, \mathrm{OUT\_OF\_RANGE})\}$
    \ENDIF
\ENDFOR
\RETURN $E$
\end{algorithmic}
\end{algorithm}

The Validate function, formalized in Algorithm~\ref{alg:validate}, performs deterministic field-level checking: for each required field it verifies presence, data type, enumeration membership, regex format including ICD-10 and CPT code structures, and numeric range, returning a detailed error set with field names and violated constraint types. Safety flags are enforced as single-value Boolean enumerations with $S.\mathrm{enum}[f] = \{\mathrm{true}\}$, so a value of false produces an INVALID\_ENUM violation.

The RepairPrompt function constructs targeted error feedback for the model. Given the original scenario, current output, error list, and schema specification, it generates a prompt that explicitly identifies violated fields, specifies the violated constraints from the schema, and requests correction while preserving clinical accuracy. This error-aware approach enables the model to focus repair efforts on specific violations rather than regenerating the entire output.

Under this procedure, every generated output including the final repair attempt is validated before a scenario is recorded as a failure. In our evaluation runs, validation of the final-iteration outputs was applied to stored model outputs during analysis; because the validator is deterministic, this is equivalent to in-loop validation and yields identical compliance outcomes.

\subsection{Evaluation Metrics}

We employ four metrics to evaluate framework effectiveness and enable cross-model comparison.

Valid@K measures the percentage of scenarios achieving schema compliance by iteration $k$. This metric captures convergence efficiency: high Valid@1 indicates most errors resolve quickly, minimizing computational overhead. We report Valid@K for $k$ from zero to three, where $k=0$ represents baseline performance without repair.

Compliance Rate computes the binary success rate as the ratio of compliant scenarios to total scenarios. This serves as the primary metric for cross-model analysis, enabling direct comparison across models.

Error Distribution provides a taxonomy of violation types. We categorize errors into four classes: INVALID\_ENUM for values outside allowed enumerations, INVALID\_FORMAT for regex pattern mismatches, TYPE\_MISMATCH for incorrect data types, and OUT\_OF\_RANGE for numeric values exceeding valid bounds; the validator additionally flags MISSING\_FIELD for absent required fields, which did not occur in our evaluation. This taxonomy reveals whether failures stem from systematic patterns or random mistakes.

Statistical Significance testing employs McNemar's exact test for paired binary compliance outcomes, appropriate for comparing schema compliance before and after repair on the same clinical scenarios. We report exact two-sided p-values, absolute improvements in percentage points, 95 percent percentile confidence intervals for the improvement computed from 100{,}000 paired bootstrap resamples at the scenario level, and the relative reduction in noncompliance.

\section{EXPERIMENTAL SETUP}

We conduct a multi-model evaluation to assess schema compliance across diverse architectures and identify whether noncompliance patterns are model-specific or systematic. Our experimental design consists of two components: a clinical scenario benchmark spanning ten medical specialties, and a multi-vendor model selection strategy that enables robust conclusions about architecture generalizability.

\subsection{Clinical Scenario Benchmark}

We constructed a benchmark of 320 clinical scenarios across ten medical specialties, each following the structured case-vignette format used in medical education and licensing examinations. Each scenario represents a typical clinical presentation documented in practice guidelines and major medical textbooks. Scenarios were designed based on cardiology guidelines from the American Heart Association and American College of Cardiology~\cite{c21}, infectious disease case definitions from the Centers for Disease Control and Prevention, and clinical presentations described in Harrison's Principles of Internal Medicine~\cite{c13} and Kumar and Clark's Clinical Medicine~\cite{c14}.

Because the purpose of this benchmark is schema compliance evaluation rather than clinical diagnostic accuracy assessment, scenarios were designed as guideline-grounded structured prompts to test interoperability requirements rather than validated patient cases requiring clinical expert review.

Each scenario includes three components: patient demographics including age and relevant medical history, chief complaint describing the primary reason for clinical encounter, and key clinical findings sufficient for structured diagnostic and treatment decision-making. This structure mirrors real-world clinical documentation while providing sufficient information for schema-compliant output generation.

The dataset distribution was informed by encounter frequencies reported in national ambulatory care surveys~\cite{c15}. Cardiology accounts for 48 scenarios at 15.0 percent, Respiratory Medicine and Infectious Disease each contain 40 scenarios at 12.5 percent, Gastroenterology and Neurology each include 32 scenarios at 10.0 percent, General Medicine comprises 36 scenarios at 11.3 percent, Endocrinology and Trauma/Orthopedics each contain 26 scenarios at 8.1 percent, and Dermatology and Psychiatry each include 20 scenarios at 6.3 percent. This distribution totals 320 scenarios providing sufficient statistical power for per-specialty analysis while maintaining practical dataset size.

\subsection{Multi-Model Evaluation Design}

We evaluate three open-source LLMs via local deployment using Ollama, selecting models from different companies to test whether schema noncompliance is architecture-specific or systematic across vendors. The selected models are Qwen2.5 7B~\cite{c25} from Alibaba Cloud representing smaller models with non-Western training emphasis, Llama 3.1 8B~\cite{c26} from Meta representing mid-size widely adopted architectures, and Gemma2 9B~\cite{c27} from Google DeepMind representing larger instruction-tuned models.

This multi-vendor selection enables robust conclusions about noncompliance patterns. Consistent failure rates across the three companies would indicate shared characteristics of medical training corpora rather than model-specific architectural limitations. Divergent patterns would suggest vendor-specific issues addressable through training improvements. The parameter range from 7B to 9B tests whether model size influences schema compliance while remaining within practical deployment constraints for local healthcare environments.

Our evaluation protocol compares two conditions for each model. The strong baseline employs explicit schema specification in the prompt, representing current best practices for structured output generation. The strong plus repair condition applies our validation-repair framework to the same prompts. We use temperature zero to ensure deterministic reproducible outputs across all experiments. Each model processes all 320 scenarios once per condition, yielding 960 model--scenario pairs.

All experiments use identical prompt templates and schema specifications to ensure fair comparison. The prompt includes the clinical scenario, explicit field requirements, schema constraints, and two example outputs demonstrating proper format, i.e., few-shot prompting with explicit schema specification. This strong baseline establishes an upper bound on what current prompting techniques can achieve without post-hoc validation and repair.

\section{RESULTS AND FINDINGS}

Our multi-model evaluation across 960 model--scenario pairs reveals three key findings. First, schema noncompliance is consistent across the three evaluated vendors despite different training approaches. Second, error taxonomy shows 96 percent of failures are representation-level format violations. Third, validation-repair achieves consistent deployment-oriented schema compliance across all architectures with practical convergence within two iterations. We present detailed analyses addressing each research question.

\subsection{RQ1: Cross-Vendor Consistency of Noncompliance}

Table~\ref{tab:performance} presents compliance rates across the three models and two evaluation conditions.

\begin{table}[t]
\caption{Multi-Model Performance Across 320 Clinical Scenarios}
\label{tab:performance}
\centering
\begin{tabular}{lccc}
\toprule
\textbf{Model} & \textbf{Strong} & \textbf{Strong+Repair} & \textbf{Improvement} \\
& \textbf{Baseline} & & \textbf{(pp)} \\
\midrule
Qwen2.5 7B & 91.6\% & 99.4\% & +7.8 \\
(Alibaba) & (293/320) & (318/320) & \\
\midrule
Llama 3.1 8B & 85.9\% & 98.4\% & +12.5 \\
(Meta) & (275/320) & (315/320) & \\
\midrule
Gemma2 9B & 89.4\% & 99.1\% & +9.7 \\
(Google) & (286/320) & (317/320) & \\
\midrule
\textbf{Overall} & \textbf{89.0\%} & \textbf{99.0\%} & \textbf{+10.0} \\
& \textbf{(854/960)} & \textbf{(950/960)} & \\
\bottomrule
\end{tabular}
\end{table}

Our multi-model evaluation addresses RQ1 by revealing that schema noncompliance is consistent across vendors. Despite different companies, training data sources, and parameter counts ranging from 7B to 9B, all three models exhibit remarkably similar baseline compliance rates spanning only 5.7 percentage points from 85.9 to 91.6 percent. This narrow variance suggests the problem stems from shared characteristics of medical training corpora, which emphasize clinical accuracy over healthcare IT standards, rather than model-specific architectural limitations.

Even the best baseline model still produces 8.4 percent noncompliant outputs, while the weakest baseline produces 14.1 percent noncompliant outputs. This error rate remains problematic for healthcare system integration where single violations cause billing rejections or workflow disruptions. Notably, the smallest model at 7B parameters achieves the highest baseline compliance at 91.6 percent, demonstrating that parameter count alone does not predict interoperability readiness.

\subsection{RQ2: Error Taxonomy and Root Causes}

Table~\ref{tab:taxonomy} presents error distributions across models and violation types. Each noncompliant output is counted once by its first detected violation; mandatory-field presence is likewise enforced, and no missing-field violations occurred in our evaluation, so the table reports the four observed error categories.

\begin{table}[t]
\caption{Error Taxonomy Before Repair}
\label{tab:taxonomy}
\centering
\begin{tabular}{lcccc}
\toprule
\textbf{Error Type} & \textbf{Qwen} & \textbf{Llama} & \textbf{Gemma} & \textbf{Total} \\
\midrule
INVALID\_ENUM & 22 & 11 & 21 & 54 (51\%) \\
INVALID\_FORMAT & 5 & 32 & 11 & 48 (45\%) \\
TYPE\_MISMATCH & 0 & 2 & 0 & 2 (2\%) \\
OUT\_OF\_RANGE & 0 & 0 & 2 & 2 (2\%) \\
\midrule
\textbf{Noncompliant Outputs} & \textbf{27} & \textbf{45} & \textbf{34} & \textbf{106} \\
\bottomrule
\end{tabular}
\end{table}

Cross-model error analysis addressing RQ2 reveals a critical insight: 96 percent of validator-detected failures are representation-level format violations. INVALID\_ENUM and INVALID\_FORMAT together account for 102 of the 106 noncompliant outputs. This pattern remains consistent across all three models with ENUM plus FORMAT comprising 100 percent for Qwen, 96 percent for Llama, and 94 percent for Gemma.

Representative error cases demonstrate models follow clinical writing conventions while violating healthcare IT standards. INVALID\_ENUM errors include using BD instead of BID, where both terms mean twice daily but the former represents international Latin convention while the latter is the required US standard for EHR systems. Case sensitivity violations such as qd versus QD also appear. INVALID\_FORMAT errors include CPT\_74140 instead of 74140, where the prefix adds semantic clarity for human readers but violates billing system requirements expecting bare numeric codes. Period formatting inconsistencies such as I21.0 versus I210 and coding system mismatches such as G0438 representing HCPCS codes instead of CPT codes also occur.
These errors reveal models trained on medical literature naturally learn international conventions and semantic annotations that improve human readability in research contexts but violate strict standardization required for healthcare system interoperability. Because clinical correctness was not independently evaluated in this study, we characterize these violations as representation-level rather than knowledge-level errors. This finding nonetheless challenges assumptions that improved medical training corpora alone will solve deployment readiness: models can produce clinically conventional outputs while remaining systematically noncompliant with interoperability standards.

\subsection{RQ3: Validation-Repair Effectiveness}

\begin{figure}[t]
\centering
\includegraphics[width=\columnwidth]{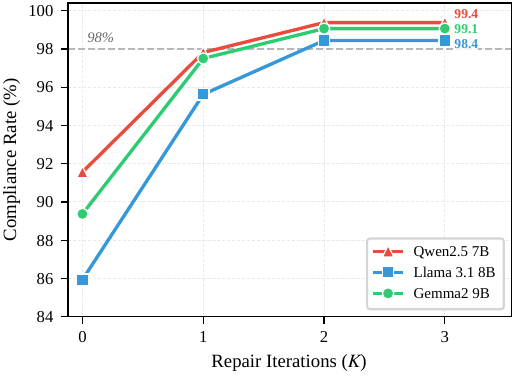}
\caption{Valid@$K$ convergence across three LLMs, computed from the cumulative convergence counts in Table~\ref{tab:convergence}. All models exceed the 98 percent threshold by iteration two, and no additional scenarios converge at the third iteration.}
\label{fig:convergence}
\end{figure}

\begin{table}[t]
\caption{Statistical Analysis of the Repair Framework}
\label{tab:significance}
\centering
\footnotesize
\setlength{\tabcolsep}{3pt}
\begin{tabular}{lcccc}
\toprule
\textbf{Model} & \textbf{Impr./Regr.} & \textbf{$\Delta$ (pp)} & \textbf{95\% CI (pp)} & \textbf{Exact $p$} \\
\midrule
Qwen2.5 7B & 25 / 0 & +7.8 & [5.0, 10.9] & $5.96{\times}10^{-8}$ \\
Llama 3.1 8B & 40 / 0 & +12.5 & [9.1, 16.2] & $1.82{\times}10^{-12}$ \\
Gemma2 9B & 31 / 0 & +9.7 & [6.6, 13.1] & $9.31{\times}10^{-10}$ \\
\bottomrule
\end{tabular}
\end{table}

\begin{table}[t]
\caption{Convergence Distribution Across Iterations}
\label{tab:convergence}
\centering
\begin{tabular}{lcccc}
\toprule
\textbf{Model} & \textbf{K=0} & \textbf{K=1} & \textbf{K=2} & \textbf{Failed} \\
& \textbf{(Baseline)} & & & \\
\midrule
Qwen2.5 7B & 91.6\% & 6.3\% & 1.6\% & 0.6\% \\
& (293) & (20) & (5) & (2) \\
\midrule
Llama 3.1 8B & 85.9\% & 9.7\% & 2.8\% & 1.6\% \\
& (275) & (31) & (9) & (5) \\
\midrule
Gemma2 9B & 89.4\% & 8.1\% & 1.6\% & 0.9\% \\
& (286) & (26) & (5) & (3) \\
\bottomrule
\end{tabular}
\end{table}

Validation-repair achieves deployment-oriented schema compliance exceeding 98 percent across all three architectures. Table~\ref{tab:significance} confirms highly significant improvements: every scenario that changed compliance status moved from noncompliant to compliant, with 25, 40, and 31 improved scenarios and zero regressions, exact McNemar p-values below $10^{-7}$, and absolute improvements ranging from 7.8 to 12.5 percentage points. These improvements correspond to relative reductions in noncompliance of 92.6, 88.9, and 91.2 percent for Qwen, Llama, and Gemma respectively, demonstrating robust evidence for framework effectiveness independent of model architecture.

Figure~\ref{fig:convergence} and Table~\ref{tab:convergence} reveal practical convergence patterns. By iteration one, an average of 97.0 percent of scenarios achieve compliance, showing that most violations are corrected after a single feedback step. By iteration two, all models reach at least 98 percent compliance with Valid@2 ranging from 98.4 to 99.4 percent, supporting downstream interoperability requirements. No additional scenarios converge at the third iteration, indicating diminishing returns beyond two repair attempts; $K_{\max}=3$ therefore serves as a conservative termination cap. The framework is also computationally light: including repair calls, it requires an average of 1.11, 1.20, and 1.14 LLM invocations per scenario for Qwen, Llama, and Gemma respectively.

Critically, convergence patterns remain consistent across the three models despite architectural differences. All models reach similar Valid@1 rates spanning 95.6 to 97.8 percent and Valid@2 rates spanning 98.4 to 99.4 percent. This cross-vendor convergence demonstrates the framework provides reliable interoperability compliance independent of model vendor or size, addressing RQ3 affirmatively with both statistical and practical validation.

\paragraph{Residual failures}
Ten model--scenario pairs remained noncompliant after bounded
repair, comprising two for Qwen2.5 7B, five for Llama 3.1 8B,
and three for Gemma2 9B. A representative residual violation
was the composite dosage unit \texttt{mg/kg/hour}, which falls
outside the restricted unit enumeration
\{\texttt{mg}, \texttt{ml}, \texttt{g}, \texttt{mcg},
\texttt{units}\} used in this study. Because compliance is
defined conjunctively across all required fields, a single
remaining field-level violation causes the complete output to
remain noncompliant even when other violations have been
corrected. These unresolved cases motivate the proposed
human-review fallback after the bounded repair loop. This
analysis concerns conformance to the study schema and does
not assess the clinical correctness of the generated content.

\subsection{Cross-Specialty Performance Analysis}

\begin{figure*}[t]
\centering
\includegraphics[width=0.82\textwidth]{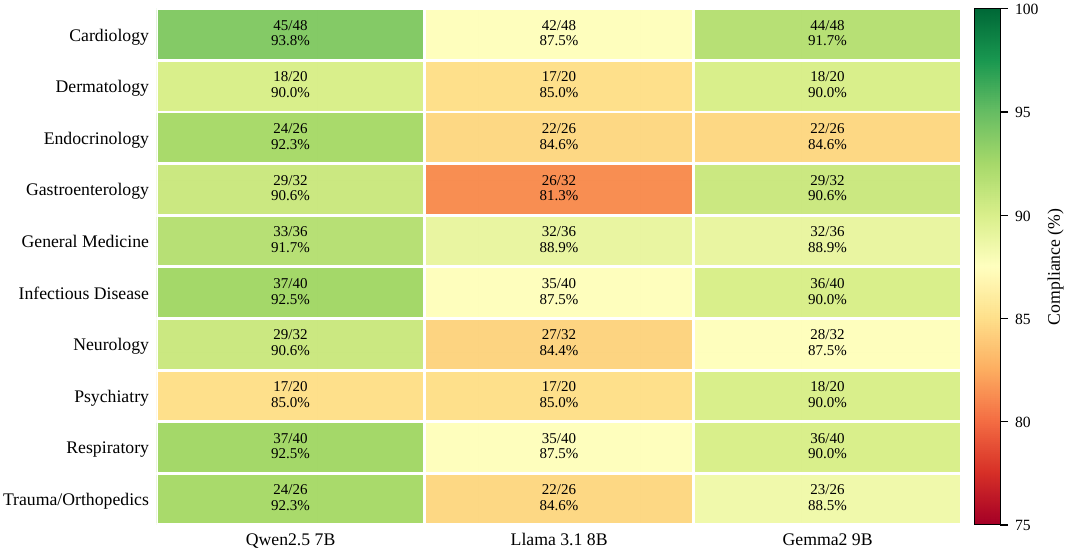}
\caption{Baseline compliance by model and medical specialty. Each cell reports compliant scenarios over specialty size together with the corresponding percentage; cell color encodes the compliance percentage. The most challenging combinations include Gastroenterology for Llama 3.1 8B at 81.3 percent and Psychiatry for Qwen2.5 7B and Llama 3.1 8B at 85.0 percent. Post-repair results not shown achieve 95 percent or higher across all combinations.}
\label{fig:specialty}
\end{figure*}

Figure~\ref{fig:specialty} shows compliance rates across ten medical specialties and three models.

Cross-specialty analysis reveals variation in baseline performance across medical domains. The most challenging specialties include Gastroenterology for Llama at 81.3 percent, Neurology for Llama at 84.4 percent, and Psychiatry for Qwen and Llama at 85.0 percent. These lower rates reflect domain-specific terminology variations and coding complexity rather than fundamental model limitations. Post-repair evaluation not shown in the figure demonstrates all model-specialty combinations achieve 95 percent or higher compliance, confirming framework robustness across diverse clinical contexts.

\section{DISCUSSION}

Our findings support schema validation-repair as an effective system-level component for healthcare AI deployment. We discuss deployment implications, training data characteristics, and threats to validity.

\subsection{Implications for Healthcare Interoperability}

\begin{figure}[t]
\centering
\includegraphics[width=\columnwidth]{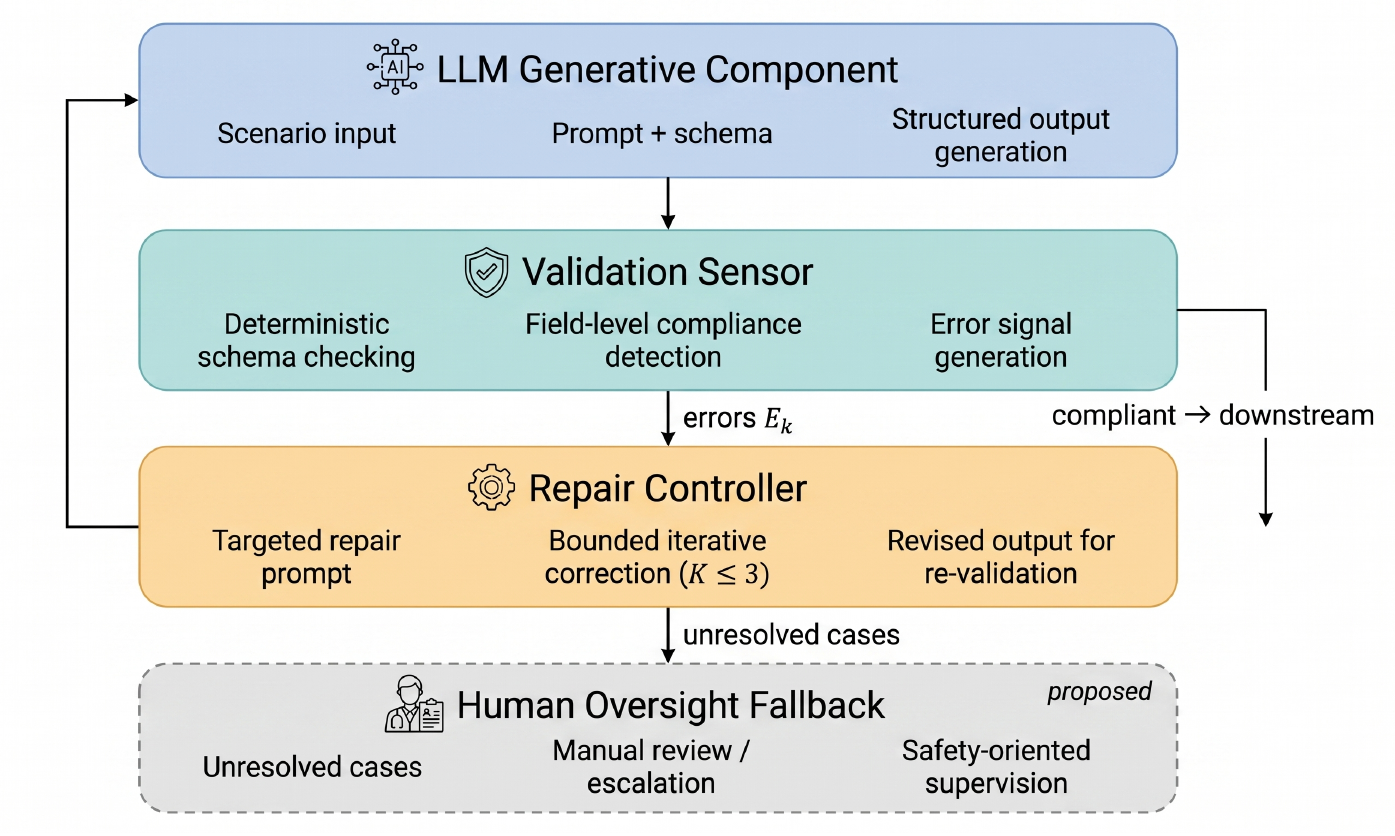}
\caption{Suggested four-layer deployment architecture for closed-loop compliance control. Compliant outputs exit after validation; the human oversight layer (dashed) is a proposed fallback for unresolved cases and was not evaluated in this study.}
\label{fig:layers}
\end{figure}

Our findings suggest that schema validation-repair should be integrated as a standard middleware layer in healthcare AI systems~\cite{c17,c19}, independent of model vendor selection: with baseline noncompliance consistent across three companies, the problem is unlikely to be solved by model choice alone. Validation-repair thus plays a role analogous to HL7 interface engines that ensure interoperability between heterogeneous clinical systems~\cite{c7}.

Figure~\ref{fig:layers} illustrates a cybernetic architecture with four functional layers: the LLM as generative component, validation as feedback sensor, repair as corrective controller with empirical convergence within two iterations, and human clinical oversight as a fallback layer handling the approximately one percent of cases that exceed automated correction. This layered architecture remains valid across model vendors, enabling LLM adoption without vendor lock-in.

Notably, the smallest model achieves the highest post-repair compliance: Qwen at 7B parameters reaches 99.4 percent, slightly exceeding Gemma2 9B at 99.1 percent. This enables lower-compute local deployment attractive for privacy-conscious healthcare settings, though regulatory compliance itself was not evaluated in this study.

\subsection{Why Closed-Loop Repair Instead of Stronger Prompting?}

A practical question is whether stronger prompting alone suffices. The strong baseline already constitutes few-shot prompting with explicit schema constraints and example outputs, yet all three models still produce noncompliant outputs at nontrivial rates, indicating a mismatch between clinical language conventions learned during training and strict interoperability requirements. Closed-loop repair addresses this by separating generation from compliance control: the LLM generates clinical content, the validator provides deterministic feedback, and the repair step performs targeted correction. Because compliance requirements are explicit and checkable, this division of responsibility is more reliable than expecting the model to internalize all schema constraints in a single pass.

\subsection{Training Data Characteristics and Implications}

The consistency of format violations across all three vendors suggests shared characteristics of medical training data: medical literature emphasizes clinical accuracy and international conventions, teaching models that BD and BID are interchangeable without distinguishing that US EHR systems require BID. Incorporating healthcare IT standards documentation such as HL7 specifications and coding guidelines into training corpora could help, but given the cross-vendor consistency of this gap, post-hoc validation-repair provides a more immediate solution that applies to any current or future model.

\subsection{Threats to Validity and Future Directions}

Schema compliance is evaluated using deterministic validators, so results depend on the selected schema constraints; our validators check code formats rather than terminology-level validity, so a well-formed but nonexistent code would not be flagged. The benchmark uses guideline-grounded structured scenarios rather than live EHR records, so results characterize schema compliance behavior under controlled interoperability tasks rather than complete deployment readiness. Finally, schema compliance does not imply clinical correctness: our study intentionally isolates interoperability compliance from diagnostic accuracy. Content-level reliability, including hallucination detection for retrieval-augmented generation~\cite{c22}, is complementary to the schema-level compliance studied here.

Our evaluation is further limited to predefined schemas rather than free-text narratives, ten specialties, three open-source models of 7B to 9B parameters, and a static schema with single-turn prompts. Future work should evaluate additional and larger models, real-world and evolving institutional schemas, conversational clinical workflows, and hybrid rule-based and LLM-based repair, as well as prompting ablations and fine-tuning with compliance-based feedback signals.

\section{CONCLUSIONS}

Through systematic multi-model evaluation across 960 model--scenario pairs, this work establishes that schema noncompliance presents a cross-vendor barrier to healthcare interoperability: three models from different vendors exhibit baseline compliance rates within 5.7 percentage points of one another, and 96 percent of failures are representation-level format violations such as international abbreviations and annotated code prefixes. This consistency suggests the problem stems from shared characteristics of medical training corpora and is unlikely to be solved by improved medical training alone.

Validation-repair achieves 99.0 percent overall compliance with practical convergence within two iterations, with statistically significant improvements of 7.8 to 12.5 percentage points across all models. Notably, the smallest model achieves the highest post-repair compliance, supporting lower-compute local deployment.

Our findings have direct implications for clinical AI deployment. Closed-loop validation-repair should be integrated as a standard system layer to support interoperability with electronic health records, independent of model vendor selection. This cybernetic approach enables healthcare organizations to adopt LLMs while maintaining schema-level compliance with FHIR-oriented, ICD-10, and CPT format requirements, transforming clinically capable but noncompliant models into interoperability-ready decision support systems suitable for downstream clinical system integration.

\addtolength{\textheight}{-9cm}   





\end{document}